\documentclass[letterpaper, 10 pt, conference]{ieeeconf}  

\IEEEoverridecommandlockouts                              

\usepackage{graphics} 
\usepackage{epsfig} 
\usepackage{mathptmx} 
\usepackage{times} 
\usepackage{amsmath} 
\usepackage{amssymb}  
\usepackage{algorithm}
\usepackage{algpseudocode}
\usepackage{bm}
\usepackage{booktabs}
\let\labelindent\relax
\usepackage{enumitem}
\usepackage{placeins}
\usepackage{hyperref}
\usepackage{xcolor}
\usepackage{multirow}
\usepackage{subfigure}
\usepackage{subcaption}
\usepackage{xcolor}
\newcommand{\numhighlight}[2]{\colorbox{#1}{#2}}

\newcommand{\numnormal}[1]{\hspace*{\fboxsep}#1\hspace*{\fboxsep}}

\definecolor{pink}{RGB}{255, 192, 203}         
\definecolor{lightpink}{RGB}{253, 235, 240}    
\definecolor{fadedpink}{RGB}{248, 200, 220}    
\definecolor{dpink}{HTML}{F28CA3}

\title{\LARGE \bf
Physics-informed Diffusion Mamba Transformer for Real-world Driving
}

\author{Hang Zhou$^{1,}$$^{2}$, Qiang Zhang$^{1}$, Peiran Liu$^{1}$, Yihao Qin$^{1}$, Zhaoxu Yan$^{2}$ and Yiding Ji$^{1}$
\thanks{$^{1}$The authors are with Robotics and Autonomous Systems Thrust, The Hong Kong University of Science and Technology(Guangzhou), Guangzhou, China
        {\tt\small hzhou269@connect.hkust-gz.edu.cn}, {\tt\small yiding\-ji@hkust-gz.edu.cn}}%
\thanks{$^{2}$The authors are with MoSense Technologies, Hong Kong SAR, China }%
}

\begin{document}

\maketitle
\thispagestyle{empty}
\pagestyle{empty}

\begin{abstract}
Autonomous driving systems demand trajectory planners that not only model the inherent uncertainty of future motions but also respect complex temporal dependencies and underlying physical laws. While diffusion-based generative models excel at capturing multi-modal distributions, they often fail to incorporate long-term sequential contexts and domain-specific physical priors. In this work, we bridge these gaps with two key innovations. First, we introduce a Diffusion Mamba Transformer architecture that embeds mamba and attention into the diffusion process, enabling more effective aggregation of sequential input contexts from sensor streams and past motion histories. Second, we design a Port-Hamiltonian Neural Network module that seamlessly integrates energy-based physical constraints into the diffusion model, thereby enhancing trajectory predictions with both consistency and interpretability. Extensive evaluations on standard autonomous driving benchmarks demonstrate that our unified framework significantly outperforms state-of-the-art baselines in predictive accuracy, physical plausibility, and robustness, thereby advancing safe and reliable motion planning.


\end{abstract}

\section{INTRODUCTION}
\label{sec:intro}
Generative diffusion model \cite{ho2020ddpm} has emerged as a powerful framework for high-fidelity data synthesis by casting sample generation as an iterative denoising process over discretized stochastic dynamics. While the U-net backbone \cite{ronneberger2015unet} is highly effective at modeling complex and multimodal distributions, it struggles to capture long-range dependencies crucial for structured outputs. The Diffusion Transformer \cite{peebles2023dit} addresses this limitation by integrating self-attention layers into each denoising step, enabling global context aggregation across spatial and temporal dimensions. This architectural enhancement improves expressivity, yielding richer sample diversity and faster convergence, while leveraging the parallelism and parameter sharing inherent in transformer models.


Despite notable progress, applying Diffusion Transformers to autonomous driving motion planning exposes several critical limitations. Most existing diffusion-based planners \cite{liao2025diffusiondrive, ajaj2023dp3, liao2024cdstraj} function as unconstrained or loosely constrained sequence generators, trained exclusively on offline datasets without explicit integration of non-holonomic vehicle kinematics, actuator constraints, or dynamic feasibility requirements. As demonstrated by DP3 \cite{ajaj2023dp3}, such models can achieve high statistical fidelity to human driving behavior yet still produce trajectories that exceed curvature or acceleration bounds when executed in closed-loop control. Some recent studies attempt to address this issue through rule-based trajectory refinement \cite{Dauner2023CORL} or post filtering \cite{meng2024diverse}. However, rule-based refinement enforces heuristic or traffic‑rule compliance, and it still does not embed vehicle dynamics or feasibility constraints into the diffusion process itself. Additionally, previous rule-based approaches operate as independent two-stage procedures rather than fundamentally constraining or optimizing the diffusion process itself, leaving the underlying generative model prone to producing physically infeasible motions.

\begin{figure}[t]
\centering
\includegraphics[ width=1.0\linewidth]{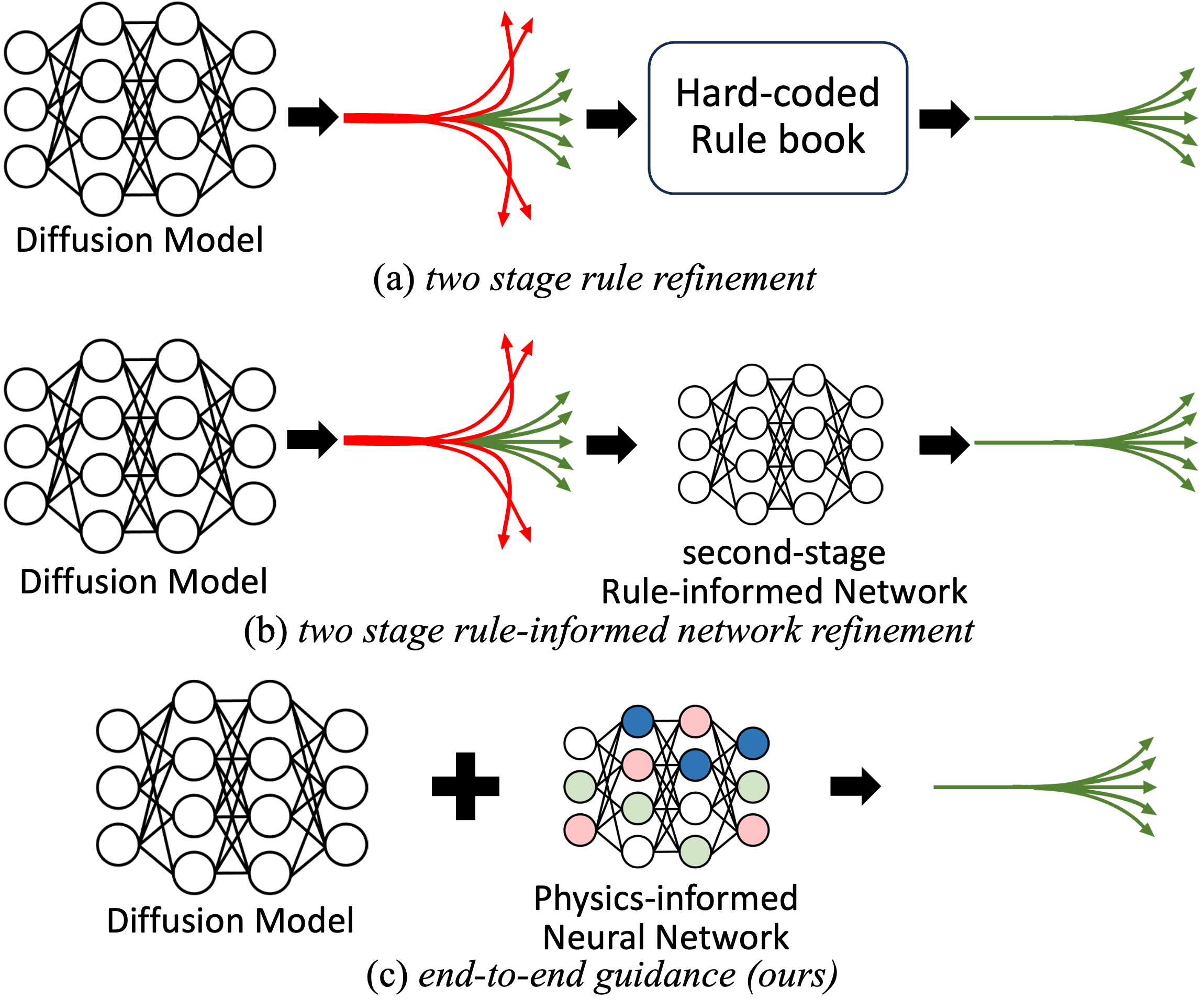}
\vspace{-10pt}
\caption{Pipelines for Diffusion model with refinement. (a) Adapts the traditional rules to filter infeasible trajectories. (b) Uses a two stage framework with a separately learned rule-informed network for trajectory refinement. (c) Our approach has a unified architecture which improves diffusion process with a physics-informed network.}
\label{fig:overview_pipeline}
\vspace{-10pt}
\end{figure}

On the other hand, although pure attention‑based architectures \cite{vaswani2017attention, dosovitskiy2021image} adept at modeling long‑range dependencies, they nonetheless lack explicit inductive biases \cite{ms-16-87-2025} that encode vehicle kinematics, collision‑avoidance constraints and the non‑Euclidean structure of dynamic road geometries. The learned representations with these built‑in priors may fail to inherently respect the physical limits of motion, the safety margins required for interaction with other road users, or the topological constraints imposed by lane connectivity and road curvature. As a result, generating physically feasible and safety‑compliant trajectories becomes particularly challenging in safety‑critical autonomous driving scenarios, where even minor violations of dynamic feasibility can lead to hazardous outcomes \cite{paden2016survey, schwarting2018planning, koschi2018representing}. Recent studies on transformer‑based trajectory planners \cite{ngiam2022scene, cui2021lookout, giuliari2021transformer} further demonstrate that, in the absence of physics‑informed constraints or domain‑specific regularization, learned policies may yield dynamically inconsistent outputs, such as abrupt curvature changes, infeasible accelerations, or unsafe proximity to obstacles, especially in dense and highly interactive traffic environments. 


The gap between expressive generative modeling and physical consistency highlights two core research challenges. First, how can a diffusion framework effectively aggregate and attend to sequential contexts from heterogeneous input streams? Second, how can trajectory predictors embed constraints to produce physically interpretable and reliable outputs? Addressing these questions is essential for bridging the divide between flexible learning-based planners and robust, physically grounded control.

To overcome these challenges, we introduce the physics-informed Diffusion Mamba Transformer. It incorporates Mamba modules immediately before each self-attention layer. Mamba’s linear-time sequence modeling with selective state spaces provides a compact, learnable summary of past diffusion contexts, dramatically reducing the quadratic complexity of full attention while preserving critical long-range dependencies. By feeding Mamba’s state representations into the subsequent attention block, the architecture gains both computational efficiency and enhanced temporal consistency. We further augment this design with a port-Hamiltonian neural network to model motion trajectories with physics information, ensuring that energy-based dynamics and conservation laws are respected throughout the generative process. Moreover, the selective state-space transitions can be guided by kinematic and collision-avoidance constraints, embedding physical priors directly into diffusion. The resulting model delivers smoother, dynamically feasible trajectories with significantly lower inference overhead, making it a compelling solution for real-time autonomous driving motion planning.

In this work, we propose a unified framework that resolves these challenges through two key innovations:
\begin{itemize}[leftmargin=*]
\item We introduce a Mamba Transformer architecture to Diffusion model and embeds attention and state-space modeling mechanisms directly into the diffusion process, enabling more effective aggregation of sequential input contexts from sensors and past motion histories.
\item We design a Port-Hamiltonian Neural Network module to seamlessly guide diffusion model with energy-based physical constraints, thereby enhancing trajectory predictions with both consistency and interpretability.

\end{itemize}

\section{RELATED WORK}
\label{sec:relatedwork}
\noindent\textbf {Diffusion Model for Autonomous Driving Planning.}
It is not long since diffusion models were first applied to autonomous driving motion planning. Early approaches lacked mechanisms to enforce high-level requirements, a gap addressed by integrating signal temporal logic into the sampling process \cite{meng2024diverse}, albeit at the expense of real-time performance. Subsequent work introduced gradient-free search \cite{yang2024diffusiones} to circumvent expensive backpropagation, but suffered from substantial sample inefficiency. A unified architecture then combined perception, prediction, and planning into a single diffusion network \cite{wang2025diffad}, simplifying deployment yet sacrificing modular interpretability. DiffusionDrive \cite{liao2025diffusiondrive} introduces diffusion models to end-to-end autonomous driving and proposes a novel truncated diffusion policy to mitigate mode collapse and the heavy computational overhead of directly adapting vanilla diffusion to traffic scenes, thereby accelerating inference while still exhibiting some behavioral bias. Recent flexible guidance scheme \cite{zheng2025diffusionplanner} enabled dynamic weighting of safety, smoothness and aggressiveness. However, it relied on manual tuning and often produced jerky trajectories.

\noindent\textbf {Attention and Mamba.}
Recent research has begun to explore the interplay between attention mechanisms \cite{vaswani2017attention} and Mamba-based architectures \cite{gu2023mamba}, revealing promising synergies for sequence modeling. Mamba, a selective state-space model, has been reformulated as a variant of linear attention Transformers, incorporating components such as input and forget gates, shortcut connections, and modified block designs to enhance expressiveness and efficiency. This reinterpretation bridges the conceptual gap between state-space models and attention-based frameworks, enabling comparative analysis and hybrid designs. For instance, Huang et al. proposed Trajectory Mamba \cite{huang2025trajectory} that integrates attention and Mamba within an encoder-decoder architecture for autonomous driving. These studies underscore the potential of combining attention with Mamba to balance long-range dependency modeling and computational efficiency. More recently, Mei et al. introduced HAMF \cite{mei2025hamf}, a hybrid framework that leverages attention mechanisms for joint scene encoding and employs Mamba-based modules for future motion decoding, achieving competitive results with a lightweight and real-time capable architecture. 

\noindent\textbf{Hamiltonian Neural Networks (HNNs)}~\cite{greydanus2019hamiltonian} are designed to learn an implicit Hamiltonian function that governs the evolution of conservative dynamical systems. In practice, however, many physical systems experience continuous energy exchange through external inputs, dissipation, or losses. To address such non‑conservative behavior, Port‑Hamiltonian Neural Networks (PHNN)~\cite{desai2021port} extend the HNN formulation by introducing explicit energy ports, enabling the modeling of energy inflow and outflow. Similarly, Dissipative Hamiltonian Neural Networks~\cite{sosanya2022dissipative} embed dissipation mechanisms to account for effects such as friction and aerodynamic drag. These extensions broaden the applicability of Hamiltonian‑based learning, offering effective tools for simulating systems where energy is not conserved. PiDT \cite{zhou2025pidt} adapts the idea of PHNN for autonomous driving and monitors energy flow with predicted action by reinforcement learning. Our approach differs as PHNN is deployed to guide trajectory output and PiDT guides action output.

\begin{figure*}[t]
\centering
\includegraphics[ width=1\linewidth]{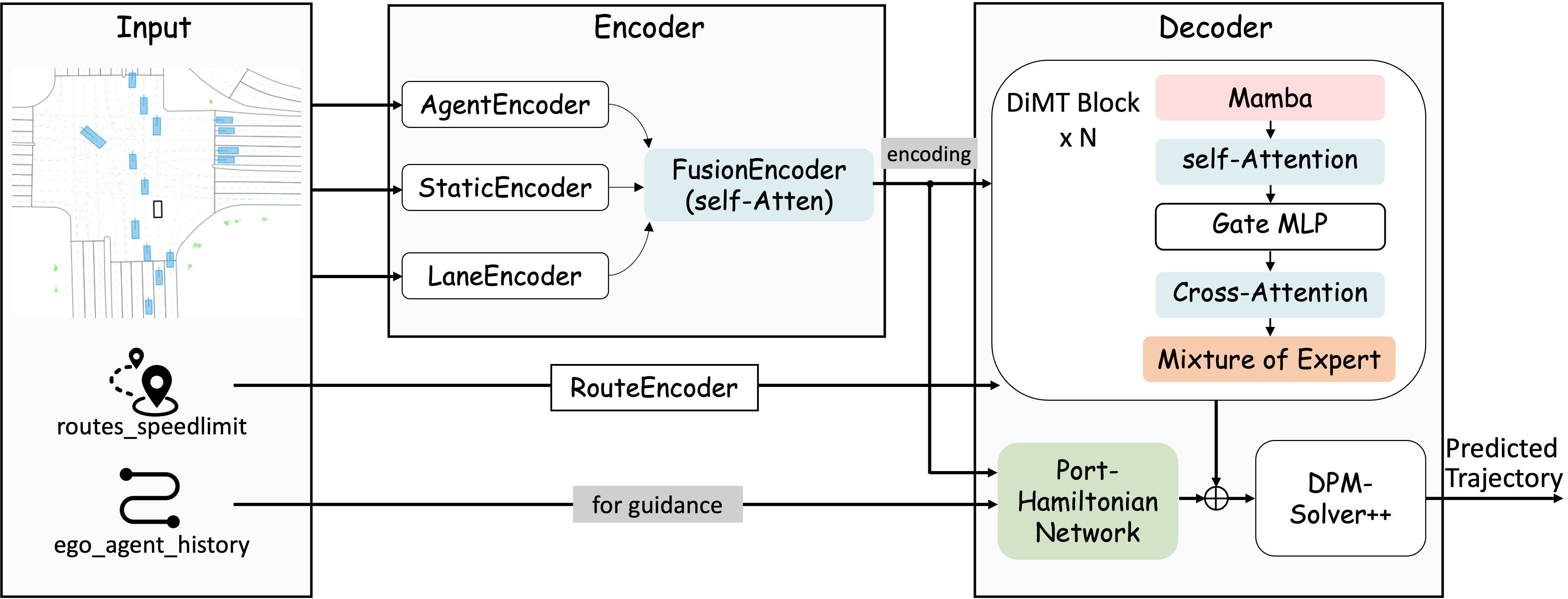}
\vspace{-10pt}
\caption{Pi-DiMT follows an encoder–fusion–diffusion design. Agent, static-object, and lane features are encoded into compact embeddings and fused via a multi-head FusionEncoder to form scene context. A DiT-based diffusion decoder, conditioned on context, route features, and the diffusion timestep, denoises trajectory tokens with the current state constrained. Finally, a Port-Hamiltonian Network provides physics-informed guidance via end-to-end training.} 
\label{fig:network_architecture}
\vspace{-10pt}
\end{figure*}

\section{PRELIMINARY}
\noindent\textbf {Port-Hamiltonian Neural Networks (PHNN)}. Classical Hamiltonian formulation presumes a closed energy‑conserving system, in which the generalized coordinates \(q\) and conjugate momenta \(p\) evolve according to  
\begin{equation}
\dot{q} = +\frac{\partial H}{\partial p}, \quad \dot{p} = -\frac{\partial H}{\partial q},
\label{eq:classical-momentum}
\end{equation}  
where \(H\) denotes the total energy of the system. Port-Hamiltonian represents non‑conservative energy systems such as propulsion or braking in autonomous systems. The momentum equation can be extended with a generalized force term  
$Q_{\mathrm{nc}} = F_{\mathrm{extern}} - F_{\mathrm{drag}}$,
where \(F_{\mathrm{extern}}\) models externally applied forces and \(F_{\mathrm{drag}}\) accounts for resistive losses. Here, $Q_{\mathrm{nc}}$
 represents the collection of non-conservative generalized forces acting on the system. The resulting momentum dynamics become  
$\dot{p} = -\frac{\partial H}{\partial q} + Q_{\mathrm{nc}}$.

A more general framework for describing systems with energy exchange is provided by the port‑Hamiltonian formalism. Defining the state vector as $x = \begin{pmatrix} q \\ p \end{pmatrix}$.
The evolution law is  
\begin{equation}
\dot{x} = \big[J(x) - R(x)\big] \nabla H(x) + G(x)u,
\end{equation}  
where \(J(x)\) is a skew‑symmetric interconnection matrix, \(R(x)\) is a symmetric positive‑semidefinite dissipation matrix, \(G(x)\) maps control inputs to the state space, and \(u\) represents external inputs such as engine or brake forces.

\section{METHOD}
In this section, we present Pi‑DiMT, a novel end‑to‑end physics‑informed diffusion model for real‑world autonomous driving motion planning. At its core is a Diffusion Mamba Transformer block that captures long‑range temporal dependencies and structured multi‑agent interactions. Pi‑DiMT also has a deigned Port-Hamiltonian Network for physics‑aware guidance to ensure kinematic and dynamic feasibility.

\subsection{Hybrid Multi‑Mechanism Diffusion Backbone for Multi‑Agent Trajectories}

We employ a diffusion transformer whose core computational unit, the Diffusion Mamba Transformer Block (DiMTBlock) integrates four complementary sequence operators within a unified residual pathway (Fig. \ref{fig:network_architecture}). First, a Mamba state‑space module enables linear‑time sequence mixing with an extended receptive field, providing efficient and expressive modelling of long‑horizon dynamics. This is followed by multi‑head self‑attention, which captures global relational dependencies between agents and across time, and a gated multilayer perceptron that refines features through dense nonlinear transformations. Cross‑attention to route and scene context reintroduces fine‑grained environmental constraints, ensuring that trajectory generation is grounded in the surrounding topology. Finally, a Mixture‑of‑Experts (MOE) \cite{jacobs1991adaptive} feedforward path allocates conditional computation capacity, allowing the model to specialise in diverse motion and interaction patterns. The fixed ordering of Mamba, Self‑Attention, Gated MLP Cross‑Attention and MoE is deliberate, progressing from temporal smoothing and latent state evolution, through global relation aggregation and feature expansion, to contextual alignment and ultimately sparse expert specialisation. LayerNorm preconditioning and residual addition are applied throughout to maintain stability and gradient flow.

A single adaptive LayerNorm‑style modulation head receives a global conditioning vector $y$ derived from agent states, scene layout, and route intent. This head generates scaling, shifting, and gating parameters for each subpath, reducing reliance on multiple parallel modulation branches and ensuring coherent semantic conditioning. Learned gates regulate residual strengths dynamically, enabling flexible path‑wise reweighting without altering the block’s structure.

Agent and context encoding is handled by tokenising the ego agent and $K$ neighbouring agents into a persistent set, encouraging implicit relational coupling and avoiding per‑agent decoding. Scene topology and route intent are jointly processed and fused with temporal embeddings to form the conditioning vector $y$. These contextual features are reintroduced during trajectory refinement through cross‑attention, where they serve as keys and values for evolving trajectory queries.

The Mixture‑of‑Experts design combines multiple shallow experts, which specialise in local motion features, with deeper experts that provide stable high‑level corrections. Per‑token gating ensures spatially and agent‑specific routing without excessive computational cost, while hybrid routing yields elastic capacity, balancing diversity and stability during both early and mature training stages. The MoE component combines top‑k routed shallow experts (adaptive specialization) with always-on deeper experts whose outputs are averaged, yielding a stabilizing backbone that mitigates early expert collapse and variance spikes typical in sparse routing. Light annealed stochastic noise on gating logits (decaying over training) supplies controlled exploration before consolidation without auxiliary load-balancing losses. This design targets a favorable trade-off between conditional capacity and optimization stability under diffusion noise perturbations.

\begin{figure}[t]
\centering
\includegraphics[width=1\linewidth]{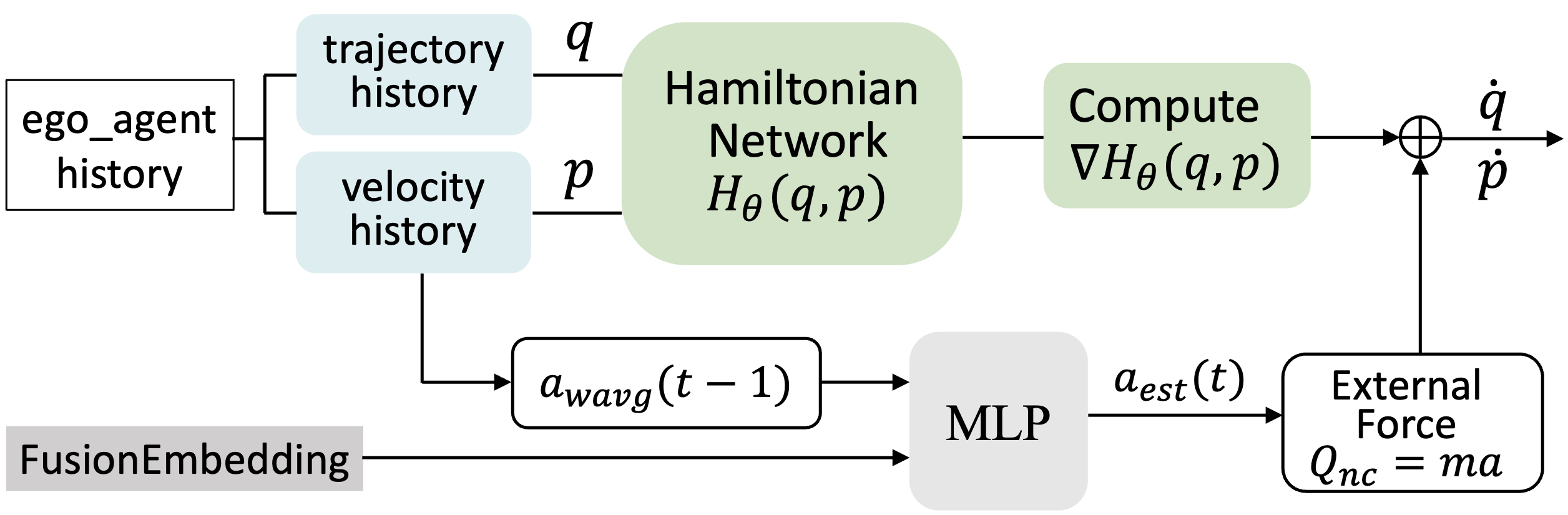}
\vspace{-5pt}
\caption{Port-Hamiltonian Network Guidance. The energy exchange is calculated by the dominant kinetic energy which is caused by acceleration. The estimated acceleration is learned by the MLP with input of weight average acceleration $a_{wavg}(t-1)$ of past 0.5s and current scenario embedding.} 
\label{fig:PHNN_architecture}
\vspace{-5pt}
\end{figure}

\subsection{Physics‑Guided Consistency via Port-Hamiltonian Guidance}
\vspace{-5pt}



In modern vehicular engineering, most notably within autonomous driving platforms, constant cruising speeds can be achieved by counteracting dissipative influences such as tire–road friction and aerodynamic drag. With these secondary losses attenuated, analytical emphasis shifts toward the energy transformations linked to changes in kinetic energy. In this simplified vehicle model, the conservative potential energy is neglected, which mathematically corresponds to assuming $\frac{\partial H}{\partial q} \approx 0$
. This assumption is justified because the vehicle is considered to move on a flat surface without position-dependent conservative forces (e.g. gravity on a slope or elastic restoring forces). As a result, the only relevant stored energy is the kinetic energy associated with the vehicle’s momentum. The dynamics reduce to 
\begin{equation}
\dot{q} = \frac{p}{m}, \quad \dot{p} = Q_{\mathrm{nc}} = F_{\mathrm{extern}} - F_{\mathrm{drag}}
\end{equation}

with the Hamiltonian given by \(H = \frac{p^{2}}{2m}\), representing purely kinetic energy.

In this regime, the principal dynamic effects arise from throttle and braking inputs, expressed through variations in acceleration. Accordingly, the non-conservative forcing term \( Q_{\text{nc}} \) is represented by the relation \( F = m \cdot a \). Within the Port-Hamiltonian Neural Network formulation, the Hamiltonian is governed primarily by the kinetic energy term, while the modeled external forcing encapsulates acceleration-driven interactions. Suppressing the contribution of lower-order dissipative mechanisms enables the framework to focus on the dominant energy flows originating from engine output and braking interventions, thereby yielding a physically coherent and computationally tractable description of the vehicle’s motion dynamics.  

\begin{equation}
a_{wavg}(t) = \frac{\sum_{i=0}^{n-1} a_{t-i}}{n}
\end{equation}

As the acceleration for the subsequent motion is unknown, we design an acceleration estimation network. It computes the moving‑average acceleration ($a_{wavg}(t)$) over the preceding 0.5 seconds and feed it together with the current scenario feature embedding, into a learnable multilayer perceptron (MLP) to estimate a physically plausible acceleration ($a_{est}$). Historical ego motion histories are decomposed into generalized coordinates \(q(t)\) (positions) and conjugate momenta \(p(t)\) (velocities).  
In the port‑Hamiltonian setting $H_\theta(q,p)$, non‑conservative effects such as propulsion and drag are incorporated via additional input channels and dissipation terms, yielding the general form  
\begin{equation}
\dot{x} = \big[J(x) - R(x)\big] \nabla H(x) + G(x)u,
\end{equation}
 
with \(x = (q,p)^\top\), skew‑symmetric interconnection matrix \(J(x)\), positive semidefinite dissipation matrix \(R(x)\), and input mapping \(G(x)\) applied to control inputs \(u\).

Guidance is performed through a small, fixed number \(S\) of symplectic‑style updates:  
\begin{equation}
\begin{aligned}
q^{(s+1)} &= q^{(s)} + \Delta t \,\frac{\partial H}{\partial p}, \\
p^{(s+1)} &= p^{(s)} - \Delta t \,\frac{\partial H}{\partial q} + Q_{\mathrm{nc}},
\end{aligned}
\quad s = 0,\dots,S-1,
\end{equation}
  
where \(Q_{\mathrm{nc}}\) represents the learned non‑conservative generalized forces calculated by learned acceleration times mass. Masks \(q_{\text{mask}}\) and \(p_{\text{mask}}\) (currently all ones) allow future extension to partial supervision or missing modalities. Only the refined \((q,p)\) corresponding to the anchored initial forecast segment are re‑injected into the leading channels of the model output, providing physics‑aligned corrections without overwriting the multimodal structure learned by the backbone. The overall PHNN architecture is designed as shown in Fig. \ref{fig:PHNN_architecture}.

\begin{table*}[t]
\footnotesize
\setlength{\tabcolsep}{1.2mm}
\centering
\begin{tabular}{
    p{2.8cm}  
    p{3.5cm} |  
    p{1.2cm}p{1.6cm}  
    p{1.2cm}p{1.6cm}  
    p{1.2cm}p{1.0cm}  
}
\specialrule{.15em}{.1em}{.1em}
\multirow{2}{*}{\textbf{Type}} &
\multirow{2}{*}{\textbf{Planner}} &
\multicolumn{2}{c}{\hspace{-4.0em}\textbf{Val14}} &
\multicolumn{2}{c}{\hspace{-4.0em}\textbf{Test14-hard}} &
\multicolumn{2}{c}{\hspace{-2.0em}\textbf{Test14}} \\
& & \textbf{NR} & \textbf{R} & \textbf{NR} & \textbf{R} & \textbf{NR} & \textbf{R} \\
\specialrule{.05em}{.05em}{.05em}
Expert & Log-replay & \numnormal{93.53} & \numnormal{80.32} & \numnormal{85.96} & \numnormal{68.80} & \numnormal{94.03} & \numnormal{75.86} \\
\hline
\multirow{6}{*}{Rule-based \& Hybrid}
& IDM \cite{Treiber2000IDM} & \numnormal{75.60} & \numnormal{77.33} & \numnormal{56.15} & \numnormal{62.26} & \numnormal{70.39} & \numnormal{74.42} \\
& PDM-Closed \cite{Dauner2023CORL}& \numnormal{92.84} & \numhighlight{pink}{\textbf{92.12}} & \numnormal{65.08} & \numnormal{75.19} & \numnormal{90.05} & \numhighlight{lightpink}{91.63} \\
& PDM-Hybrid \cite{Dauner2023CORL} & \numnormal{92.77} & \numhighlight{lightpink}{92.11} & \numnormal{65.99} & \numnormal{76.07} & \numnormal{90.10} & \numnormal{91.28} \\
& GameFormer \cite{Huang_2023_ICCV} & \numnormal{79.94} & \numnormal{79.78} & \numnormal{68.70} & \numnormal{67.05} & \numnormal{83.88} & \numnormal{82.05} \\
& PLUTO \cite{cheng2024pluto} & \numhighlight{lightpink}{92.88} & \numnormal{76.88} & \numhighlight{lightpink}{80.08} & \numnormal{76.88} & \numhighlight{lightpink}{92.23} & \numnormal{90.29} \\
& Pi-DiMT w/ refine (ours)& \numhighlight{pink}{\textbf{94.08}}  & \numnormal{91.46} & \numhighlight{pink}{\textbf{80.19}}  & \numhighlight{pink}{\textbf{81.69}}  & \numhighlight{pink}{\textbf{95.83}} & \numhighlight{pink}{\textbf{93.13}} \\
\hline
\multirow{6}{*}{Learning-based}
& PDM-Open\textbf{*}\cite{Dauner2023CORL} & \numnormal{53.53} & \numnormal{54.24} & \numnormal{33.51} & \numnormal{35.83} & \numnormal{52.81} & \numnormal{57.23} \\
& UrbanDriver \cite{scheel2021urban} & \numnormal{68.57} & \numnormal{64.11} & \numnormal{50.40} & \numnormal{49.95} & \numnormal{51.83} & \numnormal{67.15} \\
& GameFormer w/o refine \cite{Huang_2023_ICCV} & \numnormal{13.32} & \numnormal{8.69} & \numnormal{7.08} & \numnormal{6.69} & \numnormal{11.36} & \numnormal{9.31} \\
& PlanTF \cite{cheng2024rethinking} & \numnormal{84.27} & \numnormal{76.95} & \numnormal{69.70} & \numnormal{61.61} & \numnormal{85.62} & \numnormal{79.58} \\
& PLUTO w/o refine.\textbf{*}\cite{cheng2024pluto} & \numhighlight{lightpink}{88.89} & \numnormal{78.11} & \numnormal{70.03} & \numnormal{59.74} & \numhighlight{lightpink}{89.90} & \numnormal{78.62} \\
& Diffusion Planner \cite{zheng2025diffusionplanner} & \numhighlight{pink}{\textbf{89.87}} & \numhighlight{lightpink}{82.80} & \numhighlight{lightpink}{75.99} & \numhighlight{lightpink}{69.22} & \numnormal{89.19} & \numhighlight{lightpink}{82.93} \\
& Pi-DiMT (ours) & \numnormal{89.66}  & \numhighlight{pink}{\textbf{82.95}}  & \numhighlight{pink}{\textbf{77.08}}  & \numhighlight{pink}{\textbf{71.02}}  & \numhighlight{pink}{\textbf{92.45}} & \numhighlight{pink}{\textbf{87.09}} \\
\specialrule{.15em}{.1em}{.1em}
\end{tabular}
\vspace{-5pt}
\caption{\small Closed-loop planning outcomes on the nuPlan benchmark. 
Within each method category, the top-performing baseline scores are shown shaded in \textcolor{dpink}{pink}. \textcolor{fadedpink}{light pink} indicate the second-highest score. 
\textbf{*} indicates models using pre-searched reference lines as input benefit from prior knowledge. NR: non-reactive mode; R: reactive mode.}

\label{tab:benchmark}
\vspace{-5pt}
\end{table*}

During sampling, the observed current‑state slice of each agent is clamped at every DPM‑Solver \cite{MIR-2025-03-102} step via a dedicated correction hook. This anchoring prevents cumulative drift in the deterministic multistep solver, guaranteeing exact consistency with measurements at the present time while allowing the predicted future horizon to evolve stochastically. Following the diffusion backbone, the predicted trajectories are refined using a Port‑Hamiltonian neural module. 

This design yields a hybrid inductive bias: the diffusion backbone captures diverse, multimodal futures, while the Port‑Hamiltonian refinement nudges them toward dynamically consistent, physically plausible local flows. The approach offers three key advantages: (1) Hard anchoring eliminates the need for post‑hoc drift correction. (2) Low‑cost refinement imposes structure without requiring dynamics simulation in every backbone layer. (3) The latent code enables future extensions to scenario‑conditioned or uncertainty‑aware guidance.

\subsection{Continuous-Time Diffusion Training and Efficient Conditional Sampling}
Our approach formulates trajectory generation as a continuous-time variance-preserving stochastic differential equation (SDE), enabling closed-form marginal distributions. At each training step, a time variable is sampled from a uniform or alternative schedule, and analytically perturbed trajectories are used to train the network to predict either the clean signal or a scaled noise representation, depending on the chosen objective. The architecture remains consistent across settings, with only target normalization varying between modes. This continuous-time formulation avoids reliance on a fixed discrete training grid, aligns naturally with DPM-Solver++ inference, and allows the number of denoising steps to be adjusted post hoc without retraining.

\begin{table}[t]
\footnotesize
\setlength{\tabcolsep}{1.5mm}
\centering
\begin{tabular}{
    p{5.5cm} |  
    p{1.0cm}p{1.0cm}  
}
\specialrule{.15em}{.1em}{.1em}
\textbf{Planner} &
\multicolumn{2}{c}{\hspace{-2.0em}\textbf{Test14}} \\
& \textbf{NR} & \textbf{R} \\
\specialrule{.05em}{.05em}{.05em}
Diffusion Transformer & 86.64 & 78.64 \\
Diffusion Transformer + MOE \cite{scheel2021urban} & 87.92 & 81.38 \\
Diffusion Mamba Transformer + MOE & 90.89 & 85.45 \\
Diffusion Mamba Transformer + MOE + PHNN & 92.45 & 87.09 \\
\specialrule{.15em}{.1em}{.1em}
\end{tabular}
\vspace{-5pt}
\caption{\small Ablation Study. }
\label{tab:ab}
\vspace{-10pt}
\end{table}

At inference a deterministic multistep DPM-Solver++ integrator (default 10 steps, optionally 20+ for higher fidelity) advances the process. Each solver iteration (i) evaluates the conditioned model $f(x_t, t, y, \text{context})$, (ii) applies a second-order multistep update in log-SNR parameterization \cite{kingma2021vdm}, and (iii) enforces an initial\_state\_constraint that clamps the observed current slice, preventing drift of anchored states. Because training spans the continuum of $t$, increasing solver steps typically refines sharpness monotonically. The initial latent concatenates the exact observed current state with Gaussian noise over future frames (temperature scaling, e.g.\ 0.5) to preserve certainty where available and stochasticity where needed.

A joint multi-agent formulation diffuses ego and neighbor tokens simultaneously, enabling cross-agent dependencies to emerge through shared attention and Mamba mixing. Masked neighbors (padding or absent agents) are excluded via a current-state validity mask. A single conditioning vector $y$ (route encoder output plus timestep embedding) modulates all blocks, reducing parameter overhead relative to per-agent conditioning while imposing global semantic coherence.

\begin{figure*}[t]
\centering
\subfigure[right-turn cross road]{
\includegraphics[width=0.23\linewidth]{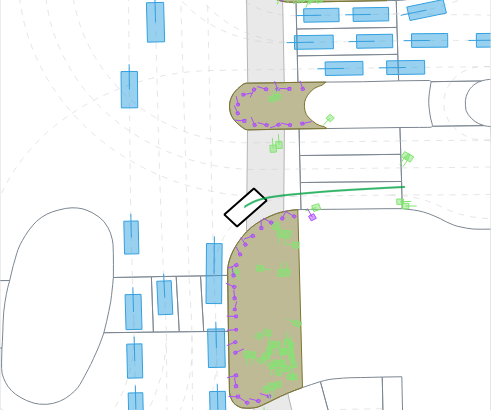}
}\hspace*{-0.1em}
\subfigure[left-turn cross road]{
\includegraphics[width=0.23\linewidth]{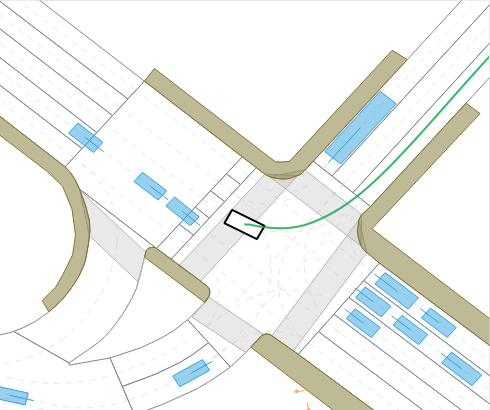}
}\hspace*{-0.1em}
\subfigure[U-turn]{
\includegraphics[width=0.23\linewidth]{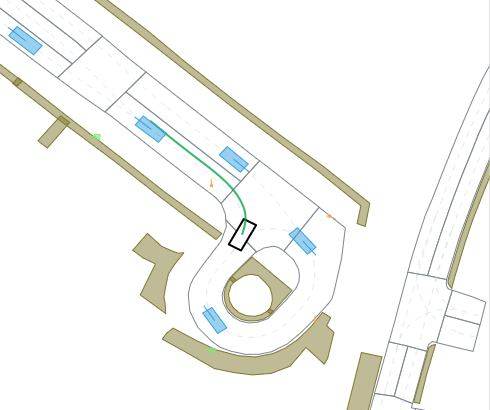}
}\hspace*{-0.1em}
\subfigure[left-turn junction]{
\includegraphics[width=0.23\linewidth]{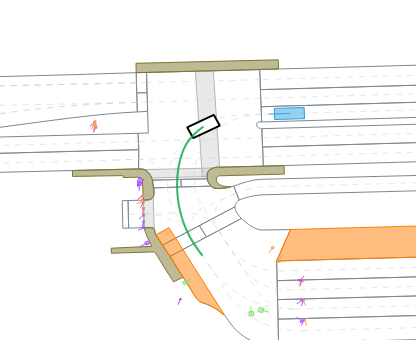}
}\\[-0.4ex]
\caption{Test Examples of Pi-DiMT completing a variety of scenarios.}
\label{fig:4}
\vspace{-10pt}
\end{figure*}

Depth scaling of the hybrid backbone is stabilized through several measures: (a) attenuating the learning rate as model depth increases, (b) using an extended warm‑up phase, (c) applying gradient norm clipping, (d) optionally reducing residual branch contributions in deeper layers, and (e) gradually annealing gating noise to prevent premature collapse of mixture‑of‑experts components. Together, these strategies mitigate the optimization challenges that arise when combining heterogeneous operators in very deep architectures.

Overall performance derives from the coordinated integration of: (1) a single modulation vector orchestrating Mamba, self-attention, MLP, and MoE; (2) physics-constrained denoising via hard anchoring plus lightweight Hamiltonian guidance (described earlier); (3) continuous-time diffusion enabling flexible solver step budgets; (4) multi-mechanism fusion (SSM + attention + MoE) within each residual block; and (5) unified multi-agent tokenization removing per-agent decoding overhead. The resulting planner is structurally expressive, physically aware, computationally adaptable, and scalable in capacity without quadratic parameter growth.

\section{EXPERIMENTAL RESULTS}
\subsection{Experimental Setup}
\noindent\textbf{Dataset, Simulator, and Metrics}. Training and experiments are conducted using the NuPlan \cite{nuplan} dataset and simulator. NuPlan comprises 1,200 hours of human driving data at 10 Hz from four cities in the US and Asia, capturing ego-vehicle states, agent trajectories, traffic signals, and high-definition maps. These are organized into roughly 15,000 multi-minute continuous logs for end-to-end motion planning research.

The nuPlan aggregate score is the per-split average of per‑scenario scores. Each scenario score starts from a weighted average of normalized continuous metrics (progress/route completion, speed limit compliance, time to collision/safe distance, lane keeping, comfort: acceleration \& jerk, efficiency) and is then gated by critical safety/rule metrics (at fault collisions, off road/drivable area violation, red light violation, wrong way, major route deviation, mission not reached). Any gating failure typically forces the scenario score to zero (or near zero). The final mean over the 14 scenarios yields the table value.

\noindent\textbf{Implementation Details}. 
We encode a structured scene with three tokenized modalities: dynamic agents (ego plus up to K neighbors), static objects, and vectorized lanes. Agent inputs consist of a temporal window of length \(V\), where each frame encodes 8-dimensional kinematic features: spatial coordinates \((x, y)\), orientation represented as \((\cos\theta, \sin\theta)\) for angular continuity, velocity components \((v_x, v_y)\), and physical dimensions \((w, l)\). Each agent is also associated with a 3-dimensional one-hot type indicator. Static objects provide a 6-dimensional feature vector \((x, y, \cos\theta, \sin\theta, w, l)\), along with a 4-dimensional categorical type code. Lane elements supply sequences of length with 8‑dimensional geometry (centerline position and local offsets to left/right boundaries) and 4‑dimensional traffic state, augmented by a scalar speed‑limit or an unknown‑limit token. Each modality is normalized, masked for missing entries, and projected through modality‑specific Mixer encoders: channel and token MLP pre‑projections, a depth stack of Mixer blocks, mean pooling, and lightweight embeddings for type, traffic, and speed‑limit metadata. Each token is assigned a unified positional–semantic representation by linearly projecting a 7-dimensional feature vector. This vector concatenates the spatial coordinates \((x, y)\), the orientation encoded as \((\cos\theta, \sin\theta)\) to ensure angular continuity, and a one-hot modality tag indicating whether the token corresponds to an agent, static object, or lane element. The concatenated token set is fused by a depth‑D self‑attention encoder with H heads and hidden feature dimension d, producing a shared latent memory for downstream diffusion. In experiments we use d = 192, H = 6, and match D to the chosen model size. 

\begin{figure*}[t]
\centering
\includegraphics[ width=1\linewidth]{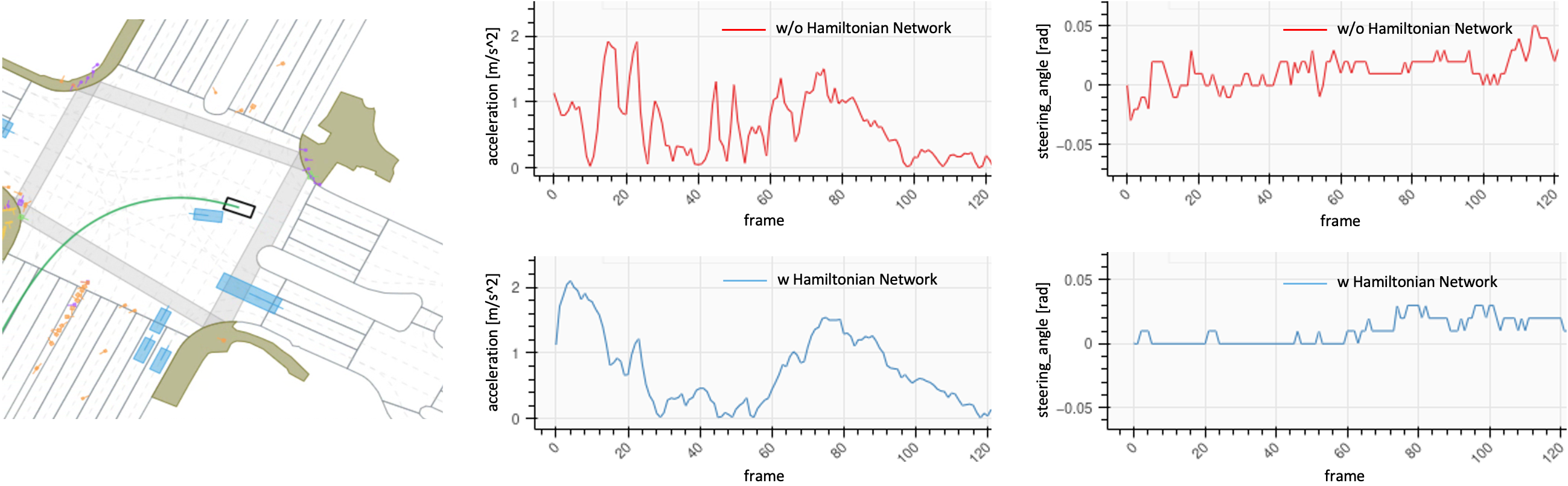}
\vspace{-10pt}
\caption{Control analysis for diffusion models with and without Port-Hamiltonian Network in left-turn cross road scenario. The PHNN‑enhanced diffusion model (blue) demonstrates substantial improvements in reducing acceleration and steering jerk.} 
\label{fig:w/o hamiltonian}
\vspace{-10pt}
\end{figure*}

\subsection{Benchmark Comparison}
Pi-DiMT is assessed in closed‑loop scenarios by benchmarking it against reactive Intelligent Driving Model (IDM) \cite{Treiber2000IDM} agents as well as non‑reactive playback agents, enabling a thorough comparison with representative rule‑based and learning‑based approaches under various traffic scenarios (Fig. \ref{fig:4}). Benchmark results are summarized in Table \ref{tab:benchmark}.

Within the learning‑based category, Pi‑DiMT delivers the strongest closed‑loop results among learning‑based planners across all splits. On Test14‑hard, it reaches 77.08 in the non‑reactive (NR) mode and 71.02 in the reactive (R) mode, outperforming Diffusion Planner by 1.09 in NR and 1.80 in R. On Test14, Pi‑DiMT attains 92.45 in NR and 87.09 in R, exceeding Diffusion Planner by 3.26 in NR and 4.16 in R, and surpassing other baselines such as PLUTO without refine and PlanTF by larger margins. On Val14, Pi‑DiMT is on par with the strongest baseline by having slightly lower in NR than Diffusion Planner (89.66 compared with 89.87) but slightly higher in R (82.95 compared with 82.80). These results indicate that our physically informed diffusion‑transformer produces trajectories that generalise well and remain executable in both evaluation modes.

Against rule‑based and hybrid planners, Pi‑DiMT is already competitive without rule refinement. On Test14, it edges PLUTO (NR: 92.45 compared with 92.23) while trailing in R (87.09 vs. 90.29), and is comparable to PDM‑Closed in NR (92.45 vs. 90.05). However, our method with rule refinement sets a new state of the art. Pi‑DiMT with refine achieves 95.83 in NR and 93.13 in R on Test14, improving over PDM‑Closed by 5.78 in NR and 1.50 in R, and over PLUTO by 3.60 in NR and 2.84 in R. On Test14‑hard, it climbs to 80.19 in NR and 81.69 in R, exceeding PDM‑Closed by 15.11 in NR and 6.50 in R. Notably, in the non‑reactive mode on Test14, it also surpasses the expert log‑replay (95.83 compared with 94.03), while maintaining a large margin in the reactive mode (93.13 compared with 75.86). These gains underscore the benefit of our physics‑informed refinement. It constrains generation toward dynamically feasible, smooth trajectories, yielding consistent advantages over heuristic rule‑based and hybrid systems in both non-reactive and reactive settings.

\subsection{Ablation Study and Analysis}

\begin{table}[t]
\footnotesize
\setlength{\tabcolsep}{1.5mm}
\centering
\begin{tabular}{
    p{5.2cm} |  
    p{1.0cm}p{1.0cm}  
}
\specialrule{.15em}{.1em}{.1em}
\textbf{Diffusion Block} &
\multicolumn{2}{c}{\hspace{-2.0em}\textbf{Test14}} \\
& \textbf{NR} & \textbf{R} \\
\specialrule{.05em}{.05em}{.05em}
Self-Atten + Cross-Atten + MOE  & 87.92 & 81.38 \\
Self-Atten + Mamba + Cross-Atten + MOE & 86.60 & 80.91 \\
Self-Atten + Cross-Atten + Mamba + MOE & 88.01 & 81.51 \\
Mamba + Self-Atten + Cross-Atten  + MOE & 90.89 & 85.45 \\
\specialrule{.15em}{.1em}{.1em}
\end{tabular}
\vspace{-5pt}
\caption{\small Ablation Study for Diffusion Block.}
\label{tab:nuplan}
\vspace{-10pt}
\end{table}


\noindent\textbf{Effect of Diffusion Mamba Transformer Block}. 
Introducing the Mamba state‑space backbone in place of the standard transformer further boosts performance to 90.89 in NR and 85.45 in R, corresponding to gains of +2.97 in NR and +4.07 in R over the MOE‑augmented Diffusion Transformer.

Table \ref{tab:nuplan} compares diffusion block configurations on the Test14 benchmark. The baseline with self‑attention, cross‑attention, and MOE module, scores 87.92 in the non‑reactive in NR mode and 81.38 in the reactive in R mode. Inserting a Mamba state‑space layer second (after self‑attention) slightly lowers performance, indicating that delaying state‑space computation weakens temporal modeling. Using Mamba after both self-attention and cross-attention gives only marginal gains. In contrast, placing Mamba first before self‑attention and cross‑attention achieves the highest scores, 90.89 in NR and 85.45 in R, improving over the baseline by +2.97 and +4.07, respectively. This ordering lets the model capture long‑range temporal structure from raw sequences via the Mamba representation before attention refines inter‑agent and contextual interactions, showing that early state‑space processing offers the strongest foundation for subsequent attention layers and yields the best closed‑loop performance.

\noindent\textbf{Effect of Port-Hamiltonian Neural Network}. 
As shown in Table \ref{tab:ab}, Augmenting the Diffusion Mamba Transformer with the proposed Port‑Hamiltonian refinement produces the highest scores: 92.45 in NR and 87.09 in R. This represents an additional improvement of +1.56 in NR and +1.64 in R over the non‑refined Mamba variant, and a total gain of +5.81 in NR and +8.45 in R relative to the baseline. Models incorporating the PHNN demonstrate markedly superior driving consistency and overall stability when compared to their non-Hamiltonian counterparts. As illustrated in Fig. \ref{fig:w/o hamiltonian}, the baseline network without Hamiltonian structure exhibits pronounced fluctuations in acceleration as well as excessive and erratic steering inputs, indicative of unstable and reactive control behavior. In contrast, the PHNN substantially mitigates these undesirable, jerky driving patterns, producing smoother and more predictable vehicle motion. This improvement can be attributed to the Hamiltonian framework’s inherent capacity to incorporate historical state information into the decision-making process, thereby enabling the model to anticipate and refine the vehicle’s future trajectory. By leveraging this temporal context, the PHNN not only reduces abrupt control variations but also enhances the robustness of the driving policy under dynamic and uncertain conditions.  

In general, PHNN processes historical vehicle dynamics to extract features that enhance the autonomous system’s tracking performance, which in turn leads to considerably smoother control inputs.

\section{CONCLUSIONS}

In summary, the proposed physics-informed Diffusion Mamba Transformer offers a physics-consistent approach to autonomous driving motion planning. The proposed DIMT architecture features Mamba module positioned before self-attention layer. It leverages linear-time sequence modeling to compress historical diffusion context and preserve essential long-range dependencies. The inclusion of a specifically designed port-Hamiltonian neural network further embeds energy-conserving dynamics into the generative process, while selective state-space transitions guided by kinematic and collision-avoidance constraints ensure adherence to physical feasibility. Together, these components produce smoother, dynamically valid trajectories with markedly reduced inference costs, establishing the approach as a promising solution for the dual demands of accuracy and efficiency in safety-critical planning tasks.  

Despite our model understanding the impact of physics and significantly improving upon it, diffusion models still lack general reasoning about the traffic world. Real-world driving involves a wide spectrum of complex, multi‑agent interactions and rare corner cases. In such situations, purely data‑driven generative processes often struggle to infer causal intent, adapt to previously unobserved scenarios, and consistently comply with traffic regulations without explicit structural reasoning. We aim to address these issues in future works.

\addtolength{\textheight}{-0cm}   





\bibliographystyle{ieeetr}
\bibliography{references}

\end{document}